\documentclass[conference]{IEEEtran}
\IEEEoverridecommandlockouts
\usepackage{cite}
\usepackage{amsmath,amssymb,amsfonts}
\usepackage{algorithmic}
\usepackage[dvipdfmx]{graphicx}
\usepackage{textcomp}
\usepackage{xcolor}
\usepackage{bm}
\usepackage{multirow}
 \UseRawInputEncoding
 
\def\BibTeX{{\rm B\kern-.05em{\sc i\kern-.025em b}\kern-.08em
    T\kern-.1667em\lower.7ex\hbox{E}\kern-.125emX}}
\begin{document}

\title{Imitation learning for variable speed motion \\generation over multiple actions\\}

\author{\IEEEauthorblockN{1\textsuperscript{st} Yuki Saigusa}
\IEEEauthorblockA{\textit{Faculty of Engineering, Information and Systems} \\
\textit{University of Tsukuba}\\
Ibaraki, Japan \\
Email: s2020741@s.tsukuba.ac.jp}
\and
\IEEEauthorblockN{2\textsuperscript{nd} Ayumu Sasagawa}
\IEEEauthorblockA{\textit{Department of Electrical and Electronic Systems} \\
\textit{Saitama University}\\
Saitama, Japan \\
Email: sasagawa.997@ms.saitama-u.ac.jp}
\and
\IEEEauthorblockN{3\textsuperscript{rd} Sho Sakaino}
\IEEEauthorblockA{\textit{Department of Intelligent Interaction Technologies} \\
\textit{University of Tsukuba}\\
Ibaraki, Japan \\
Email: sakaino@iit.tsukuba.ac.jp}
\and
\IEEEauthorblockN{4\textsuperscript{th} Toshiaki Tsuji}
\IEEEauthorblockA{\textit{Department of Electrical and Electronic Systems} \\
\textit{Saitama University}\\
Saitama, Japan \\
Email: tsuji@ees.saitama-u.ac.jp}
}

\maketitle

\begin{abstract}
Robotic motion generation methods using machine learning have been studied. Bilateral control-based imitation learning can imitate human motions using force information. Using this method, variable speed motion generation that considers physical phenomena such as the inertia and friction can be achieved. However, the previous study  focused on a simple reciprocating motion. To  learn the complex relationship between the force and speed more accurately, it is necessary to learn multiple actions using many joints. In this paper, we propose a variable speed motion generation method for multiple motions. We considered four types of neural network models for the motion generation and determined the best model for multiple motions at variable speeds. Subsequently, we used the best model to evaluate the reproducibility of the task completion time for the input completion time command. The results revealed that the proposed method could change the task completion time according to the specified completion time command in multiple motions. 
\end{abstract}


%%%%%%%%%%%%%%%%%%%%%%%%%%%%%%%%%%%%%%%%%%%%%%%%%%%%%%%%%%%%%%%%%%%%%%%%%%%%%%%%
\section{INTRODUCTION}
\label{sec:sec1}

It is expected that robots will be able to replace human laborers. Recently, studies on robotic motion generation~\cite{c29}, contact object estimation~\cite{c30}, and system identification~\cite{c31} have been reported for robotic automation. In particular, many methods to generate motions with machine learning have been studied~\cite{c1,c2,c3}. Motion generation methods based on machine learning can be divided into two approaches.

The first approach is based on reinforcement learning~\cite{c4}. In this approach, robots learn autonomously through trial and error based on the rewards that are designed by humans. The method of Levine~{\it et al}. succeeded in gripping various objects using reinforcement learning~\cite{c5}. However, the search space for robots to acquire skills is extremely large. Therefore, this approach has the disadvantage of requiring a huge number of trials and making it difficult to determine the reward~\cite{c6}.

The second approach is known as imitation learning~\cite{c8,c9,c10}. Imitation learning is supervised learning from motion data that are demonstrated by humans. Therefore, the search space is appropriately restricted and it is not necessary to design rewards. 
Recently, imitation learning using force information has been studied~\cite{c15,c16}. A robot is more adaptable to environmental changes when using force information~\cite{c17}.

As an imitation learning method using force information, we previously proposed bilateral control-based imitation learning~\cite{c20}. Bilateral control is a teleoperation technique in which two robots are used: a primary and a replica. Bilateral control synchronizes the positions of the two robots and presents the reaction force caused by the contact of the replica with the environment to the primary~\cite{c19,c32}. Bilateral control-based imitation learning can acquire human skills to compensate for control delays and dynamic interactions between the robot and the environment. In \cite{c20}, we showed that this method can achieve the same speed as a human, which has not been possible in the other imitation learning methods~\cite{c8,c9,c10,c15,c16,c17}. 

In imitation learning, generalization performance concerning operating speed has also been studied. Tani~{\it et al}. proposed a method called recurrent neural network with parametric bias, in which multiple speed behaviors are realized in a single learning model~\cite{c27}. However, the method in~\cite{c27} cannot achieve the desired operating speed because the speed information is not input during learning. To execute the motion at the desired speed, it is necessary to consider the nonlinear relationship between the force and speed (e.g. friction and inertia). To achieve this, the speed and motion information need to be provided as inputs and their association should be learned. Taking advantage of bilateral control-based imitation learning, we proposed a method that can generate variable speed motion~\cite{c21}. In~\cite{c21}, the frequency command and motion information were input into the NN model. Thus, the NN model learned the physical phenomena relating to the motion speed and achieved variable speed operation in tasks that were greatly affected by friction and inertia. However, the work in~\cite{c21} only considered reciprocating motion in which one axis moves significantly. Therefore, the basal movement was always constant, the movement space was small, and the posture almost did not change. That is, no significant change occurred in the spatial direction. The physical phenomena relating to the speed vary substantially depending on the robot posture, which has multiple Degree-of-Freedom~(DOF) mechanisms. Therefore, it is necessary to learn various speeds for multiple actions that cause large changes in the spatial direction. Furthermore, in~\cite{c21}, the motion speed was determined based on the fast Fourier transform. Thus, this method was limited to periodic motion.

In this paper, we propose a method that can generate variable speed motion for multiple actions. The proposed method can be applied to most tasks because the completion time of the task is input as a speed command to relax the requirement of periodic operation. Furthermore, a structure that is suitable for variable speed operations for multiple actions is revealed. In this study, we first examined four different models with various input layers of the completion time and task commands. We also decided on a model that can generate variable speeds for multiple actions. Thereafter, using the model, we examined the reproducibility of the task completion time for the input completion time commands.  From the results, the proposed model can change the task completion time according to the input completion time command.
The advantages of the proposed method are as follows:
\begin{itemize}
\item The NN model can learn the relationship between multiple actions and the speed.
\item The NN model for learning the variable speed operation can be applied to most time-dependent tasks.
\end{itemize}
The remainder of this paper is organized as follows. Section~\ref{sec:sec2} presents the robot control system and the bilateral control used in this study. In Section~\ref{sec:sec3}, we describe bilateral control based imitation learning and the proposed learning method. Section~\ref{sec:sec4} provides the experimental description, results, and discussion. Finally, in Section~\ref{sec:sec6}, we conclude this study and discuss future research topics.
 \begin{figure}[tb]
    \centering
        \includegraphics[width=30mm]{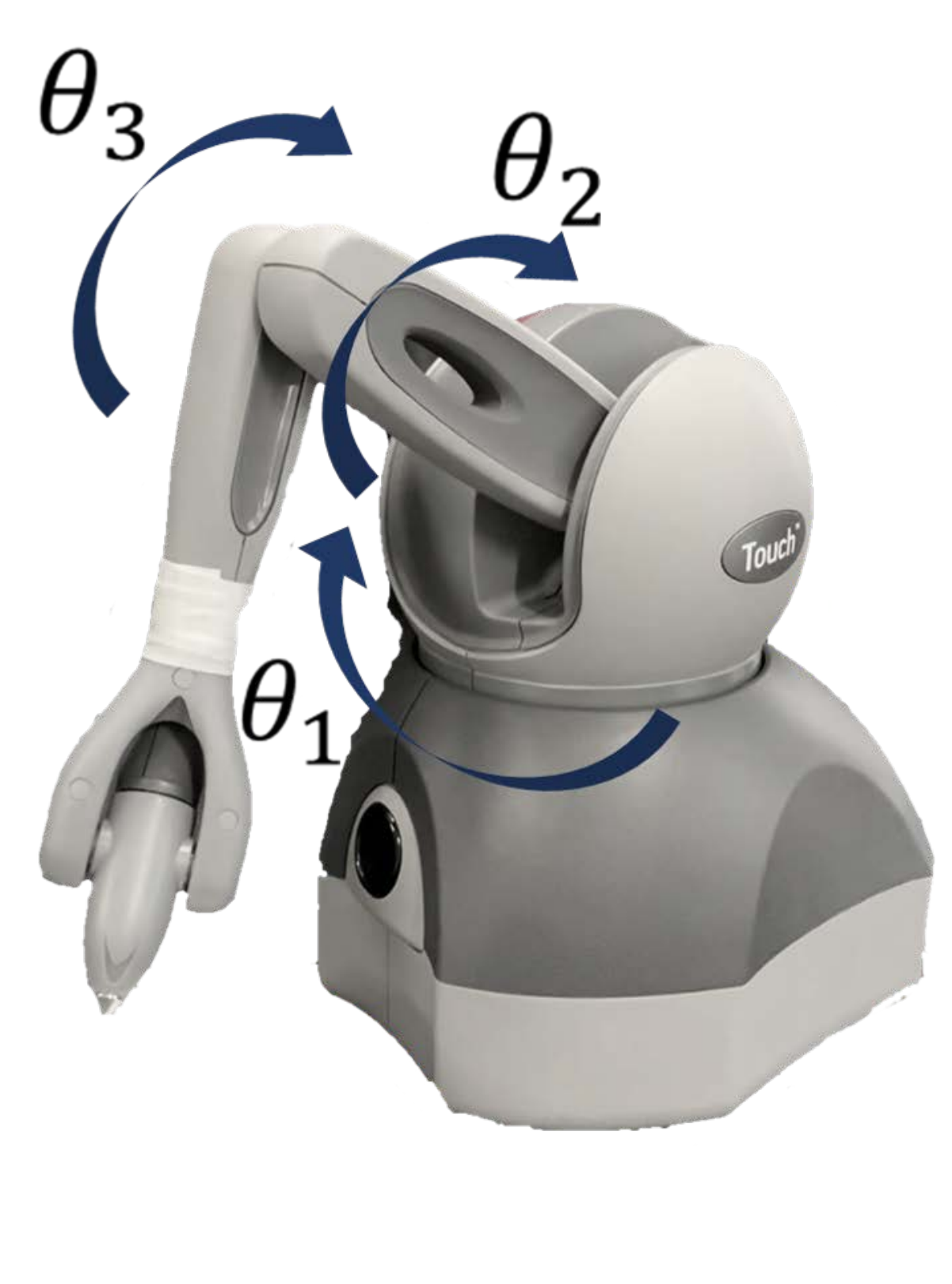}
        \caption{Manipulator~(Touch$^{\rm TM}$)}
        \label{fig:fig2}
\end{figure}

 \begin{figure}[tb]
    \centering
        \includegraphics[width=65mm]{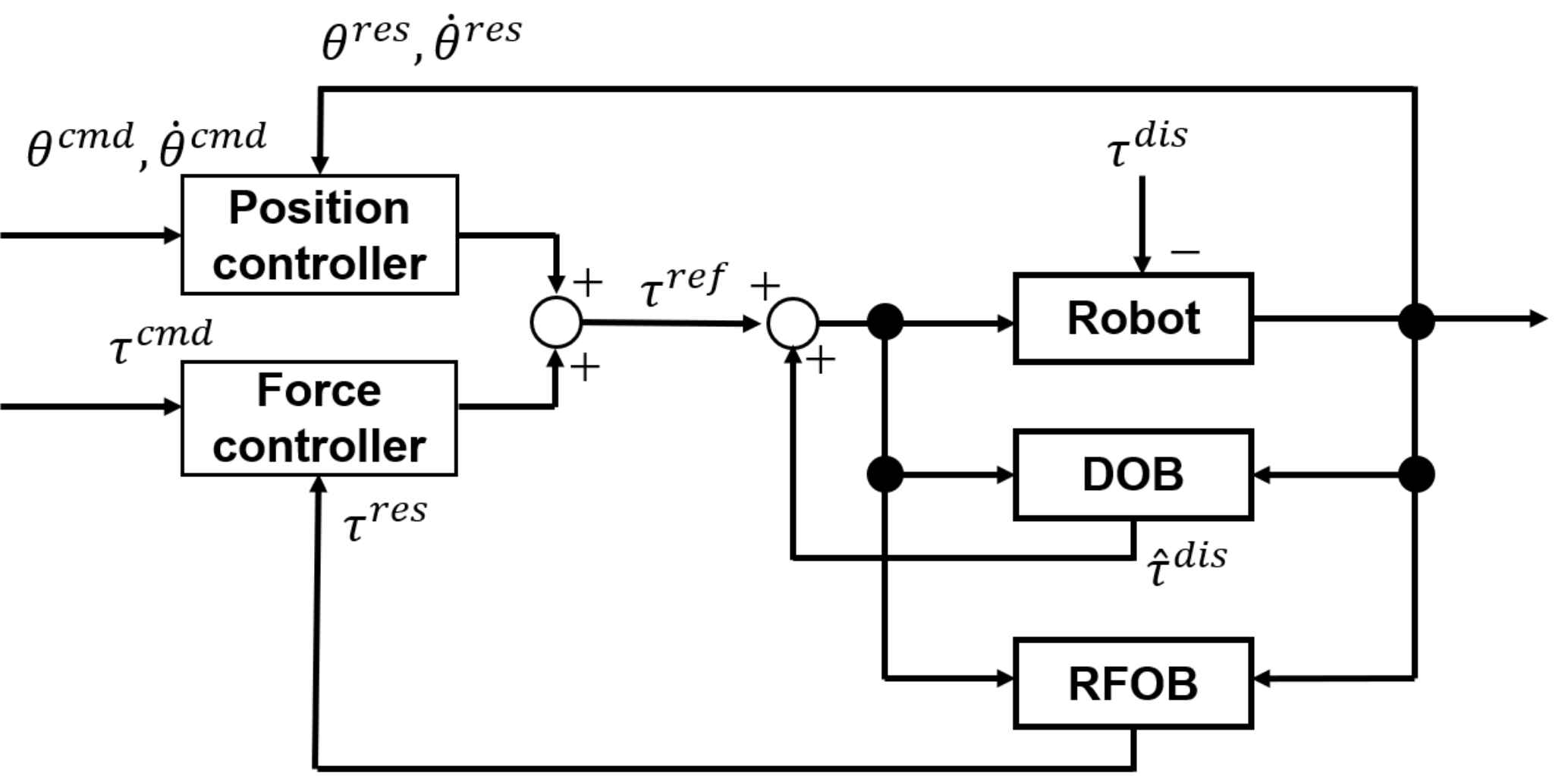}
        \caption{Block diagram of manipulator controller}
        \label{fig:fig3}
\end{figure}
\section{CONTROL SYSTEM}
\label{sec:sec2}
\subsection{Manipulator}
\label{subsec1:subsec1}
In this study, we used two Touch$^{\rm TM}$ manipulators manufactured by 3D Systems, as illustrated in Fig.~\ref{fig:fig2}.
This robot has three-DOFs and the angles $\theta_{1}$, $\theta_{2}$, and $\theta_{3}$ corresponding to each joint are defined as per Fig.~\ref{fig:fig2}. 
\subsection{Controller}
\label{subsec1:subsec2}
In this study, the manipulator control system consisted of a position controller and a force controller. The position controller consisted of a proportional and differential controller, whereas the force controller consisted of a proportional controller. The control system is depicted in Fig.~\ref{fig:fig3}. In the figure, $\theta$, $\dot{\theta}$, and $\tau$ represent the joint angle, angular velocity, and torque, respectively, and the superscripts $cmd$, $res$, $ref$, and $dis$ represent the command, response, reference, and disturbance values, respectively. The joint angle of each joint was obtained by the robot encoder and the angular velocity was calculated by its pseudo-differential. The disturbance torque $\tau^{dis}$ was calculated using a disturbance observer~(DOB)~\cite{c22} and the torque response value $\tau^{res}$ was calculated using a reaction force observer~(RFOB)~\cite{c23}. In this study, the dynamics model of the robot was assumed to be the same as that in~\cite{c20}.
\subsection{Four-channel bilateral control}
\label{subsec1:subsec3}
Bilateral control is described in this section. Four-channel bilateral control has a structure with a position controller and a force controller that are implemented on two robots: a primary and a replica. This method is effective for imitation learning using force information~\cite{c20}. Therefore, in this study, we used four-channel bilateral control. The control goal of bilateral control at each joint is represented by the following equations:
\begin{eqnarray}
  \theta^{res}_p - \theta^{res}_r = 0, \\
  \tau^{res}_p + \tau^{res}_r = 0,
\end{eqnarray}
where the subscript $p$ represents the primary and $r$ represents the replica. Furthermore, the torque reference values for the bilateral control were calculated using the following equations:
\begin{eqnarray}
   \tau^{ref}_{p} & = & - \frac{J}{2}(K_{p} + K_{d}s) (\theta^{res}_{p} - \theta^{res}_{r}) \nonumber \\ &&- \frac{1}{2}K_{f} (\tau^{res}_{p}+\tau^{res}_{r}), \label{eq3} \\
    \tau^{ref}_{r} & = &   \frac{J}{2}(K_{p} + K_{d}s) (\theta^{res}_{p} - \theta^{res}_{r}) \nonumber \\ &&- \frac{1}{2}K_{f} (\tau^{res}_{p}+\tau^{res}_{r}), \label{eq4}
\end{eqnarray}
where $K_p$ is the position control gain, $K_d$ is the velocity control gain, $K_f$ is the force control gain, $J$ is the inertia, and $s$ is the Laplace operator. In this study, $K_p = 121.0$, $K_d=22.0$, and $K_f=1.0$ were used to set each gain. 
\section{METHOD}
\label{sec:sec3}
In this section, we  explain the bilateral control-based imitation learning and describe the four types of NN models that were considered in this study.
\begin{figure}[t]
    \centering
        \includegraphics[width=65mm]{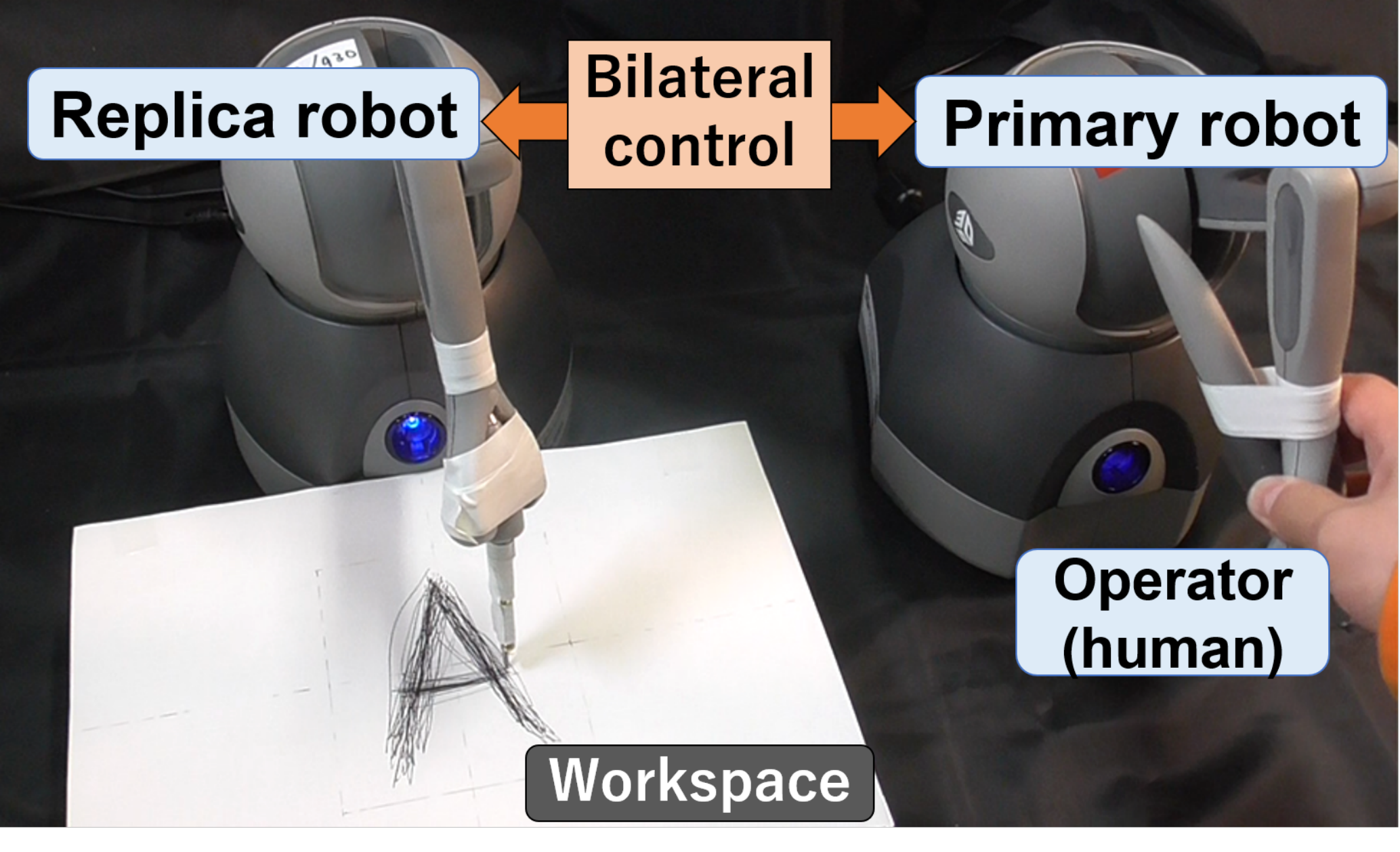}
        \caption{Collection of training data using bilateral control}
        \label{fig:fig1}
\end{figure}

 \begin{figure}[tb]
    \centering
        \includegraphics[width=80mm]{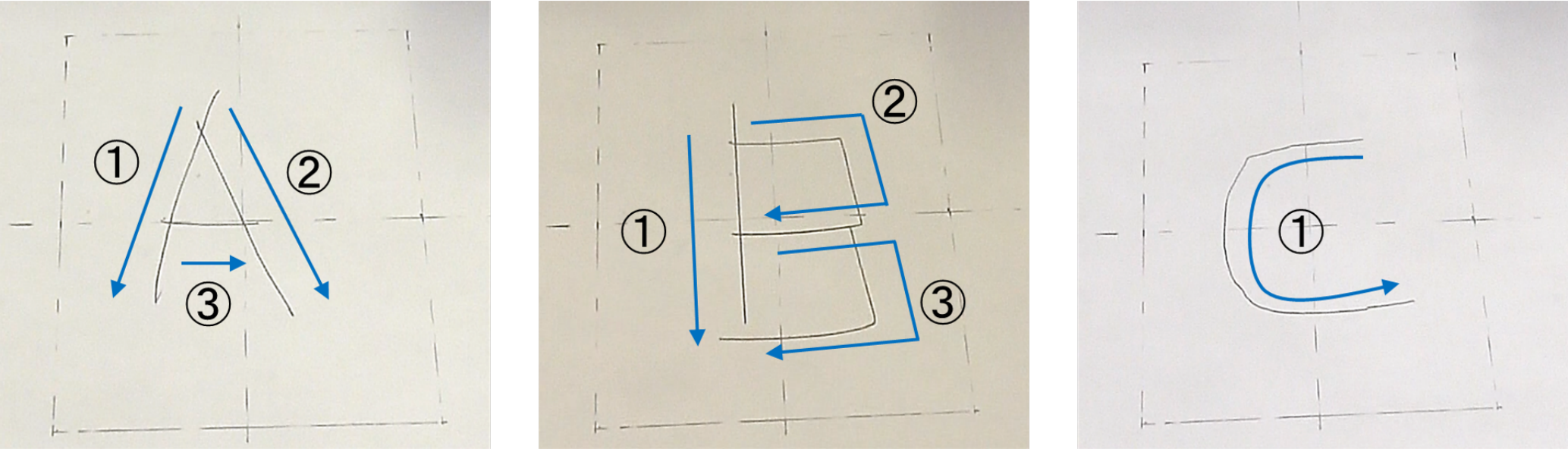}
        \caption{The stroke orders of ``A,'' ``B,'' and ``C''}
        \label{fig:fig10}
 \end{figure}
 
 \begin{table*}[tb]
\tiny
\begin{center}
\caption{Trained/untrained completion time commands and task commands used in experiment}
\begin{tabular}{c|cccccc|cccccc|cccccc}
\hline
Task command & \multicolumn{6}{c|}{A} & \multicolumn{6}{c|}{B} & \multicolumn{6}{c}{C} \\ \hline
Trained completion time command {[}s{]} & \multicolumn{2}{c}{3.00} & \multicolumn{2}{c}{6.00} & \multicolumn{2}{c|}{9.00} & \multicolumn{2}{c}{4.00} & \multicolumn{2}{c}{7.00} & \multicolumn{2}{c|}{10.00} & \multicolumn{2}{c}{2.00} & \multicolumn{2}{c}{5.00} & \multicolumn{2}{c}{8.00} \\ \hline
Untrained completion time command {[}s{]} & 2.00 & 4.00 & 5.00 & 7.00 & 8.00 & 10.00 & 2.00 & 3.00 & 5.00 & 6.00 & 8.00 & 9.00 & 3.00 & 4.00 & 6.00 & 7.00 & 9.00 & 10.00 \\ \hline
\end{tabular}
\label{tab:tab3}
\end{center}
\end{table*}
\subsection{Bilateral control-based imitation learning}
\label{subsec3:subsec1}
\subsubsection{Collection of training data}
\label{subsybsec:subsubsec1}
We experimented with the task of writing three letters ``A,'' ``B,'' and ``C'' at multiple speeds.
The training data were collected using two robots, as illustrated in Fig.~\ref{fig:fig1}.
The operator controlled the primary and executed the task by controlling the replica in the workspace. The joint angle, angular velocity, and torque of the primary and replica were saved at 1~kHz. The stroke orders of ``A,'' ``B,'' and ``C'' are depicted in Fig.~\ref{fig:fig10}. The training data were collected with a combination of the completion time and task commands, as indicated in the upper part of Table~\ref{tab:tab3}. The training data were collected 10 times for each combination. Therefore, the number of training data collected was 90 (3[task commands] $\times$ 3[completion time commands] $\times$ 10[times]). 
\subsubsection{Training the NN model}
\label{subsybsec:subsubsec2}
\begin{figure}[tb]
    \centering
        \includegraphics[width=80mm]{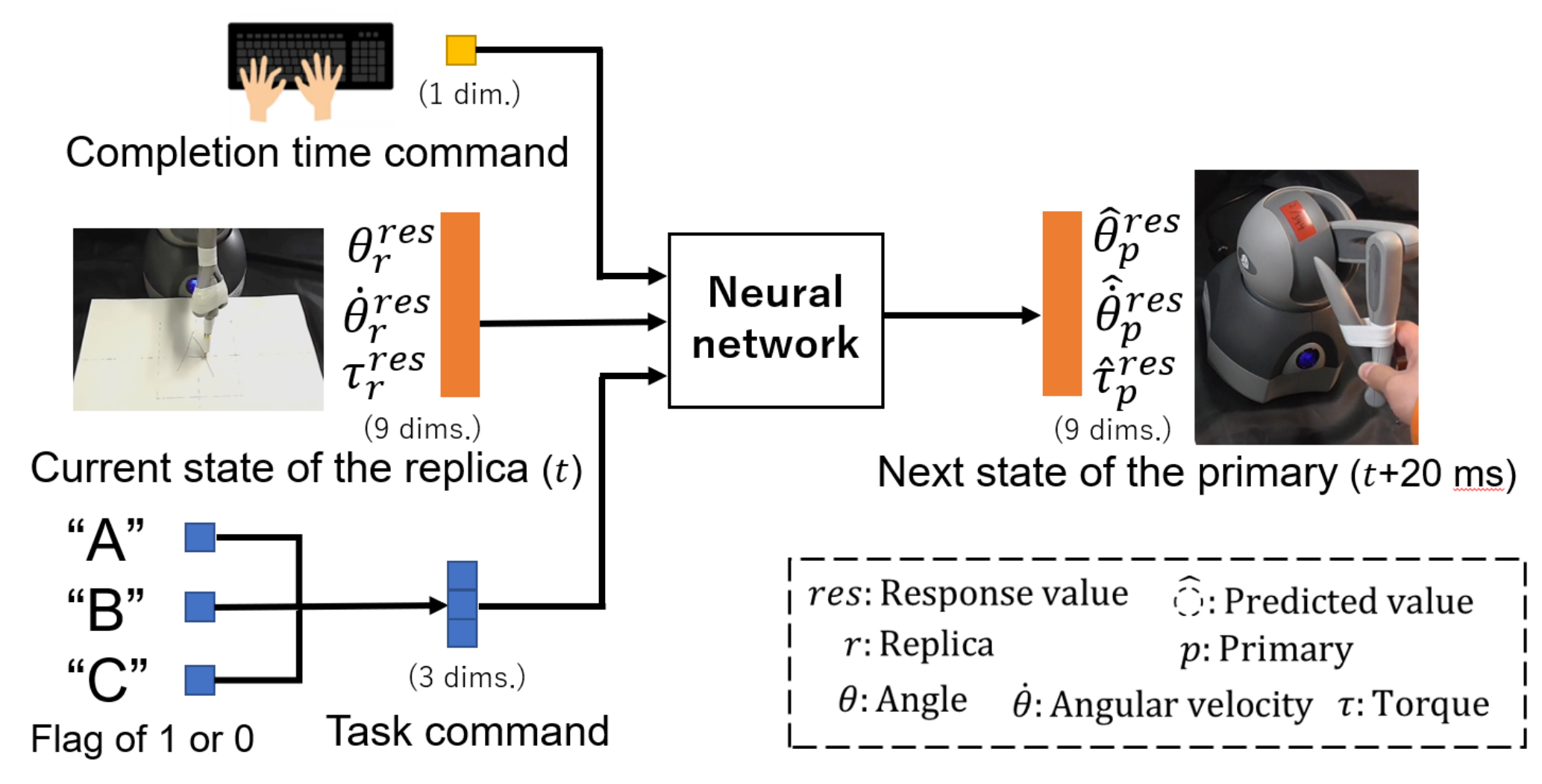}
        \caption{Outline of NN model that learns multiple actions with variable speed}
        \label{fig:fig7}
\end{figure}

An outline of the NN model used in this study is presented in Fig.~\ref{fig:fig7}. The NN model was configured with the current response value of the replica, completion time command, and task command as input, and the response value of the primary after 20~ms as output.
For the loss function, we used the mean squared error. All input values were normalized using min-max normalization. The mini-batch consisted of 100 random sets.
\subsubsection{Autonomous operation}
\label{subsybsec:subsubsec3}
The replica performed autonomous operation by using the learned model.
During the autonomous operation, the current response values of the replica were measured in real-time. These values were input into the NN model along with the completion time and task commands. The output value of the NN model was input to the replica controller as the following command value according to Eq. (4). This allowed the NN model to take the place of the primary robot and operator, thus reproducing bilateral control.
\subsection{Configuration of NN model}
\label{subsec3:subsec2}
\begin{figure}[tb]
    \centering
        \includegraphics[width=70mm]{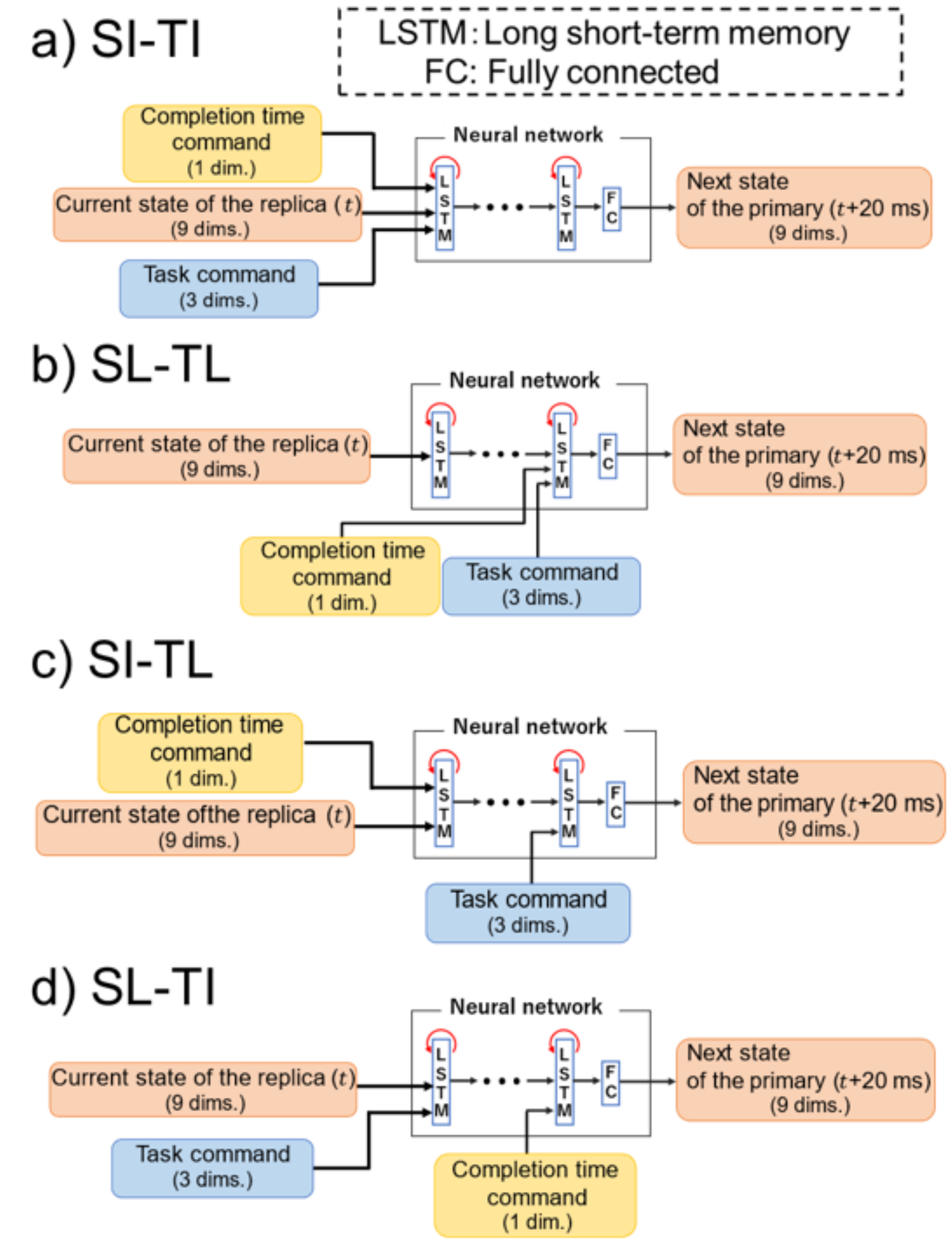}
        \caption{Configuration of four models for verification.}
        \label{fig:fig6}
\end{figure}

In this section, we describe the details of the four NN models considered in this study.
Conventionally, the NN model for learning variable speed motion has not been studied. In the field of multimodal learning, the discussion on how to fuse the information input into the NN model often arises~\cite{c26}. Therefore, inputting all information into the input layer is not always the best means of training NN models. The model must be constructed based on this point. In this study, four types of NN models with different ways of inputting the completion time and task commands were considered. The four models are presented in Fig.~\ref{fig:fig6}. For the NN model, we used long short-term memory (LSTM), in which time-series information can be learned. All models were constructed with eight LSTM layers and a fully connected layer. The nodes in the middle layer were set to 50. The descriptions of the four models are as follows:
\begin{itemize}
 \item[a)] SI-TI: All information is input into the input layer.
 \item[b)] SL-TL: The completion time command and task commands are input into the last LSTM layer.
 \item[c)] SI-TL: The completion time command is input into the input layer and the task command is input into the last LSTM layer.
 \item[d)] SL-TI: The task command is input into the input layer and the completion time command is input into the last LSTM layer.
\end{itemize}
\section{EXPERIMENT}
\label{sec:sec4}
We conducted preliminary experiments to determine the best NN model for generating multiple actions at variable speeds. Thereafter, further experiments were conducted to validate the effectiveness of the best NN model. This section describes the experiments and provides a detailed discussion of the results.
\subsection{Comparison of NN models}
\label{subsec4:subsec1}
\subsubsection{Description}
\label{subsubsec4:subsec1}
A letter-writing task was performed using the four models described in Section~\ref{subsec3:subsec2}. The success of the task was defined as writing the letters in the same stroke order as in Fig.~\ref{fig:fig10} without stopping. Furthermore, humans should be able to recognize the letters. To succeed in this task, it is necessary to consider the complex relationship between the speed and force, as well as the relationships between movements. In particular, at low speeds, the frictional force becomes large and it is necessary to apply force to prevent the stroke from stopping. By comparing the success rates of the tasks, we determined the best model for generating variable speed movements for multiple actions.
Each model was trained for 12,000 iterations using the training data described in Section~\ref{subsybsec:subsubsec2}.
In this experiment, evaluations were conducted for all combinations of the three task commands and the nine completion time commands illustrated in Table~\ref{tab:tab3}. Five trials were conducted for each condition. Therefore, 135 (3[task commands] $\times$ 9[completion time commands] $\times$ 5[times]) trials were conducted using each model.
\subsubsection{Results}
\label{subsubsec2:subsubsec1}
The experimental results are displayed in Table~\ref{tab:tab5}. SI-TL, in which the completion time command was input in the input layer and the task command was input in the last LSTM layer, exhibited the highest success rate. SI-TL could succeed in the task even when the unlearned combination of the completion time and task commands were input.
The overall success rate of the other models was less than 30\%.
Therefore, SI-TL was the best model for learning the variable speed motion over multiple actions.
\subsubsection{Discussion}
\label{subsubsec2:subsubsec2}
For SL-TL and SL-TI, a longer completion time resulted in a lower success rate. In these models, the completion time command was not input into the input layer. Therefore, the relationship between the friction and the motion speed could not be learned correctly and it was not possible to operate with an appropriate force against friction. In NNs, the information with a high level of abstraction appears in the layer close to the inputs, and vice versa. Therefore, the task completion time, which is information with a high level of abstraction, should be provided in the input.

SI-TI had many mistakes in writing ``B,'' and ``C'' confusedly. SI-TI inputs all of the information into the input layer. Because the task command was highly task-specific information, the low correlation between the layer close to the input and the task command would make the learning difficult. If the amount of training data is increased and the number of epochs is increased, the task success rate of SI-TI may be improved. In fact, results exceeding SI-TL were sometimes found when the number of epochs was set to 30000 or more. However, collecting a large amount of training data and spending a lot of time for training is very costly. Therefore, it is desirable to be able to learn efficiently with a small amount of training data.

SI-TL can learn efficiently with less training data. In SI-TL, highly abstracted information; that is, the completion time command, was provided in the input, whereas task-specific information; that is, the task command, was provided close to the output. This structure matches the architecture of NNs, which produce the highest level of abstraction at the input layer and task-oriented abstraction at the output layer. Thus, to learn variable speed motion generation for multiple motions efficiently, it is necessary to reflect the task commands after the appropriate extraction of the dynamic information.

The results showed that the success rate of "B" was low. Recently, a learning method called autoregressive learning has been proposed. Autoregressive learning improves learning efficiency and enables action generation that is robust to changes in the environment~\cite{c29}. Therefore, adapting it to the SI-TL model is expected to enable more complex actions and more efficient learning.
\begin{table*}[tb]
\begin{center}
\caption{Task execution results of each model}
\begin{tabular}{|c|c|c|c|c|c|c|c|c|c|c|c|}
\hline
\multirow{3}{*}{Model} & \multirow{3}{*}{\begin{tabular}[c]{@{}c@{}}Task \\ Command\end{tabular}} & \multicolumn{10}{c|}{Success Rate {[}\%{]}} \\ \cline{3-12} 
 &  & \multicolumn{9}{c|}{Speed Command {[}sec{]}} & \multirow{2}{*}{Total} \\ \cline{3-11}
 &  & 2.00 & 3.00 & 4.00 & 5.00 & 6.00 & 7.00 & 8.00 & 9.00 & 10.00 &  \\ \hline
\multirow{3}{*}{SI-TI} & A & 20(1/5) & 20(1/5) & 20(1/5) & 0(0/5) & 0(0/5) & 0(0/5) & 0(0/5) & 0(0/5) & 0(0/5) & \multirow{3}{*}{16(21/135)} \\ \cline{2-11}
 & B & 0(0/5) & 0(0/5) & 0(0/5) & 0(0/5) & 0(0/5) & 0(0/5) & 0(0/5) & 0(0/5) & 0(0/5) &  \\ \cline{2-11}
 & C & \textbf{80(4/5)} & \textbf{80(4/5)} & \textbf{100(5/5)} & \textbf{100(5/5)} & 0(0/5) & 0(0/5) & 0(0/5) & 0(0/5) & 0(0/5) &  \\ \hline
\multirow{3}{*}{SL-TL} & A & 0(0/5) & \textbf{100(5/5)} & 40(2/5) & 0(0/5) & 0(0/5) & 0(0/5) & 0(0/5) & 0(0/5) & 0(0/5) & \multirow{3}{*}{24(32/135)} \\ \cline{2-11}
 & B & 0(0/5) & 0(0/5) & 0(0/5) & 0(0/5) & 0(0/5) & 0(0/5) & 0(0/5) & 0(0/5) & 0(0/5) &  \\ \cline{2-11}
 & C & \textbf{100(5/5)} & \textbf{100(5/5)} & \textbf{100(5/5)} & \textbf{100(5/5)} & \textbf{100(5/5)} & 0(0/5) & 0(0/5) & 0(0/5) & 0(0/5) &  \\ \hline
\multirow{3}{*}{SI-TL} & A & 0(0/5) & \textbf{100(5/5)} & \textbf{100(5/5)} & \textbf{100(5/5)} & \textbf{100(5/5)} & 0(0/5) & \textbf{100(5/5)} & \textbf{100(5/5)} & \textbf{100(5/5)} & \multirow{3}{*}{\textbf{74(100/135)}} \\ \cline{2-11}
 & B & 0(0/5) & 0(0/5) & \textbf{100(5/5)} & 0(0/5) & 0(0/5) & 0(0/5) & \textbf{100(5/5)} & \textbf{100(5/5)} & \textbf{100(5/5)} &  \\ \cline{2-11}
 & C & \textbf{100(5/5)} & \textbf{100(5/5)} & \textbf{100(5/5)} & \textbf{100(5/5)} & \textbf{100(5/5)} & \textbf{100(5/5)} & \textbf{100(5/5)} & \textbf{100(5/5)} & \textbf{100(5/5)} &  \\ \hline
\multirow{3}{*}{SL-TI} & A & 40(2/5) & 60(3/5) & \textbf{100(5/5)} & \textbf{100(5/5)} & 40(2/5) & 0(0/5) & 0(0/5) & 0(0/5) & 0(0/5) & \multirow{3}{*}{27(37/135)} \\ \cline{2-11}
 & B & 20(1/5) & 60(3/5) & 0(0/5) & 0(0/5) & 0(0/5) & 0(0/5) & 0(0/5) & 0(0/5) & 0(0/5) &  \\ \cline{2-11}
 & C & \textbf{100(5/5)} & \textbf{100(5/5)} & \textbf{100(5/5)} & 20(1/5) & 0(0/5) & 0(0/5) & 0(0/5) & 0(0/5) & 0(0/5) &  \\ \hline
\end{tabular}
\label{tab:tab5}
\end{center}
\end{table*}
 \begin{figure*}[tb]
    \centering
        \includegraphics[width=140mm]{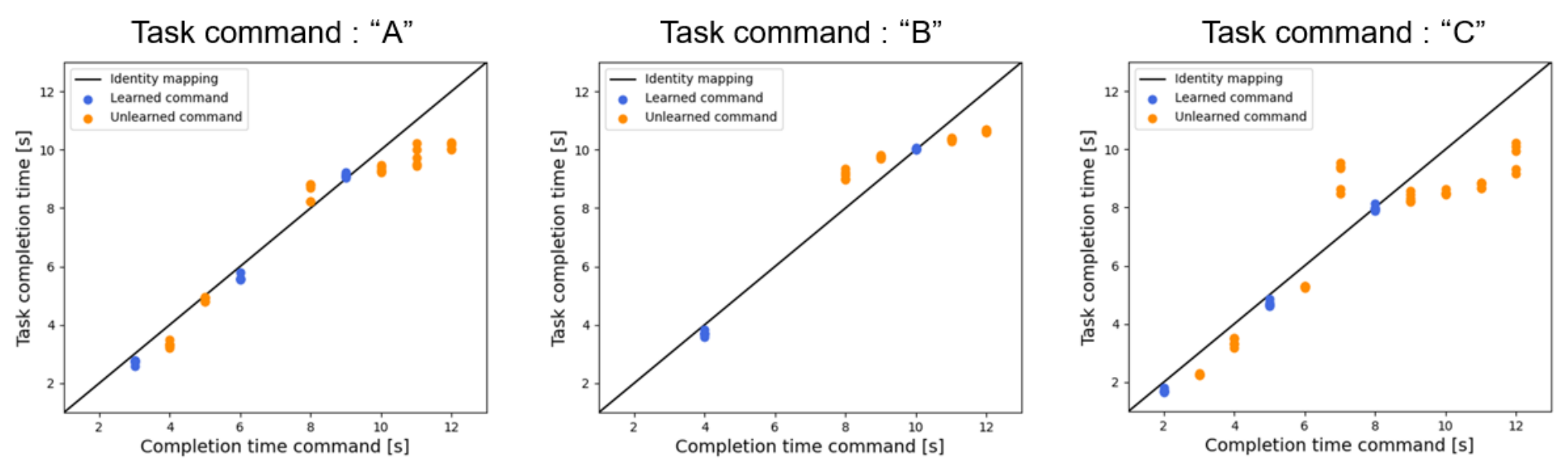}
        \caption{Completion time command and actual task completion time of SI-TL}
        \label{fig:fig12}
\end{figure*}
\subsection{Examination of task completion time repeatability for completion time command}
\label{subsec4:subsec2}
\subsubsection{Description}
\label{subsubsec4:subsec2}
We investigated the reproducibility of the task completion time for the input completion time command of SI-TL. By comparing the input completion time command with the actual task completion time, we could examine whether SI-TL could learn the relationship between each action and speed.
\subsubsection{Results}
\label{subsubsec3:subsubsec1}
The experimental results are presented in Fig.~\ref{fig:fig12}. In this figure, only the successful completion of the task is plotted. There are few plot points for ``B'' because there were many failures. In Fig.~\ref{fig:fig12}, the black line represents the identity mapping. When the plots are along the line, this indicates that the reproducibility of the completion time is ideal.
SI-TL could not perfectly reproduce the task completion time according to the completion time command. However, in most cases, the task was completed in a time close to the completion time command. 
As an additional experiment, three extrapolation completion times were tested. When the completion time command of 1.00 was input, the task could not be completed. However, when the 11.00 or 12.00 completion time commands were entered, the success rate of the task was 100\%.
Furthermore, for completion time commands of 11.00 and 12.00, SI-TL could slow down the operation by inputting slower completion time commands.
\subsubsection{Discussion}
\label{subsubsec2:subsubsec2}
\begin{figure}[tb]
    \centering
        \includegraphics[width=80mm]{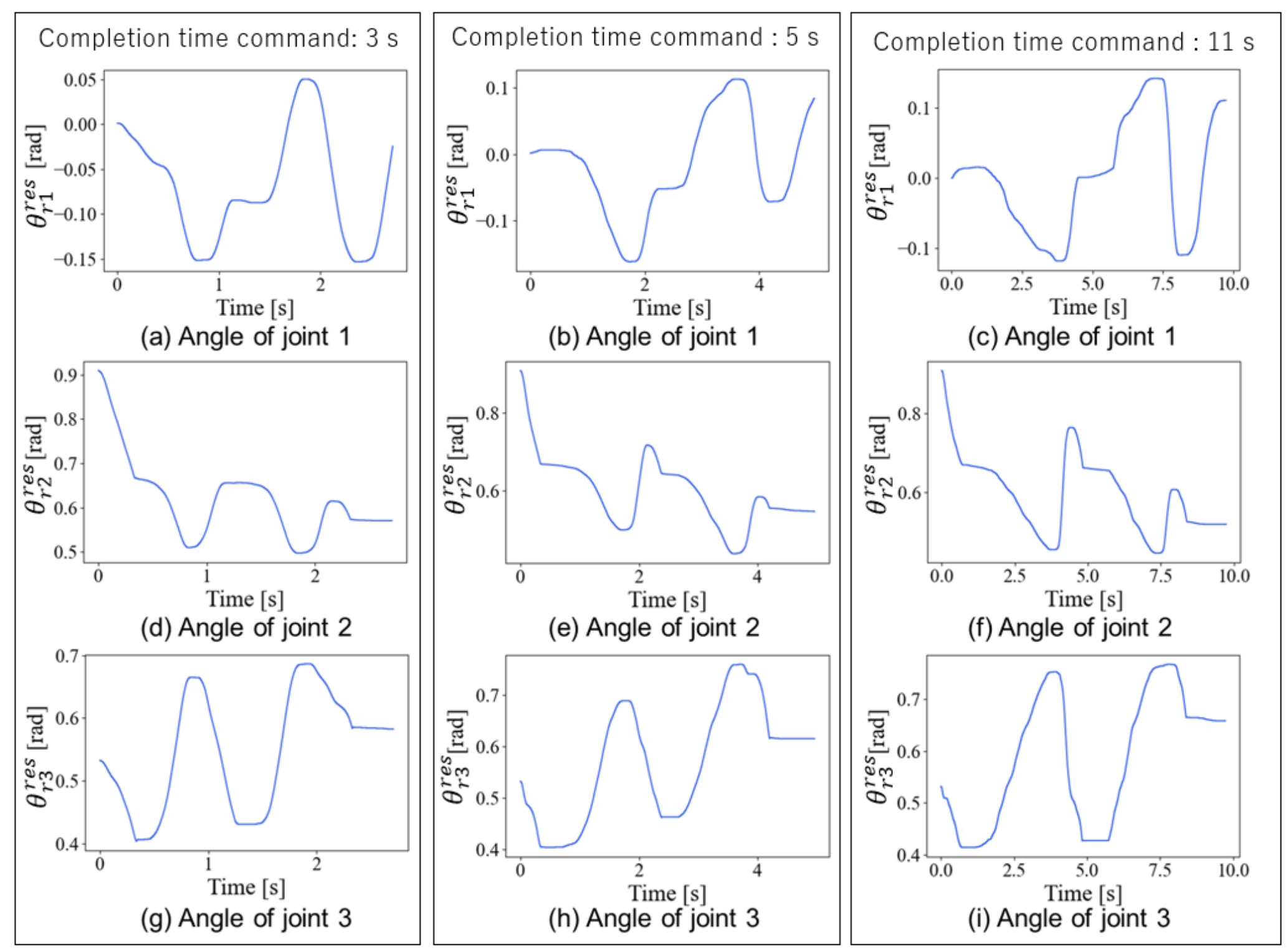}
        \caption{Angle response values during autonomous operation}
        \label{fig:fig13}
\end{figure}

\begin{figure}[tb]
    \centering
        \includegraphics[width=80mm]{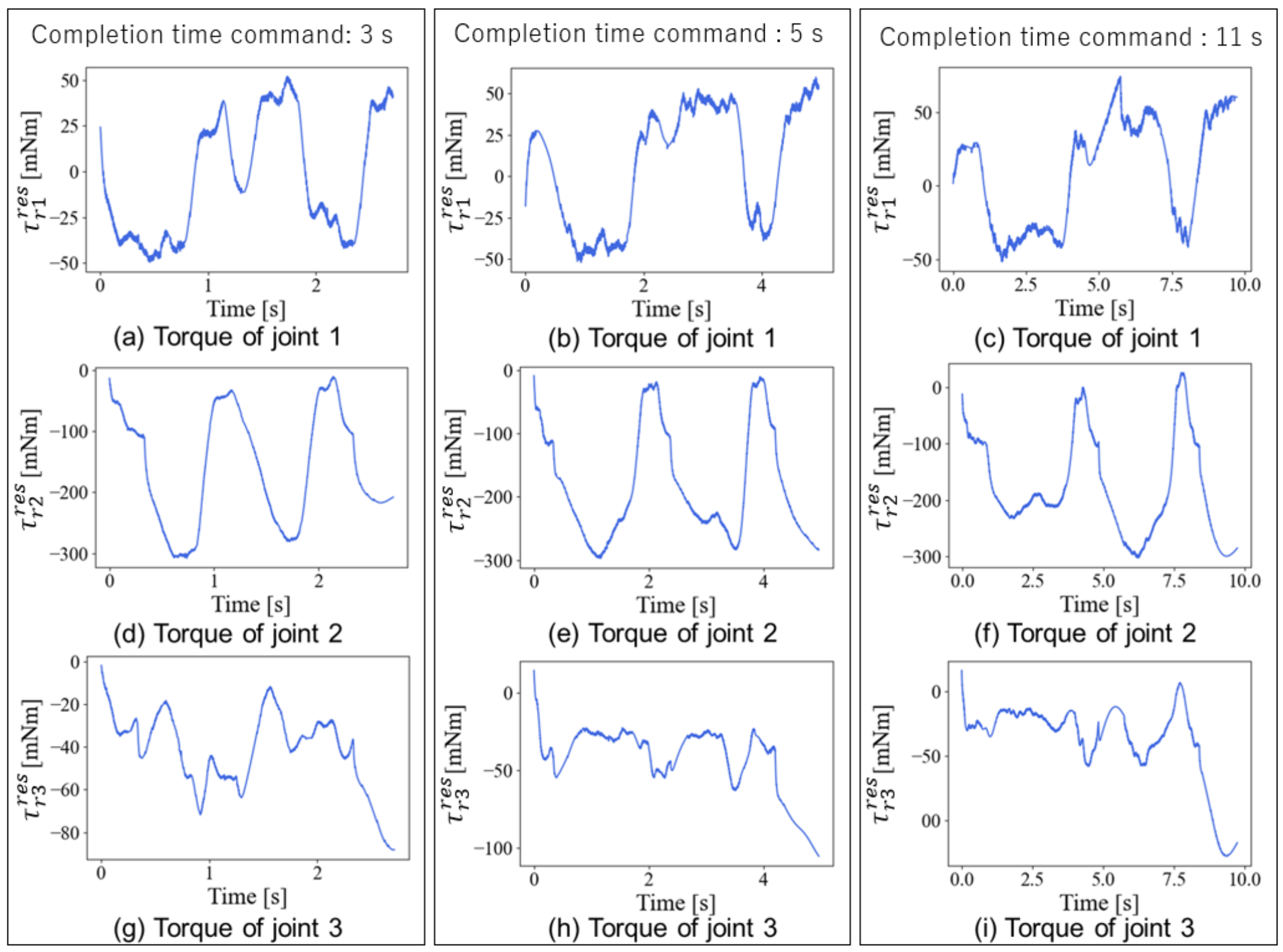}
        \caption{Torque response values during autonomous operation}
        \label{fig:fig14}
\end{figure}

The angular and torque response values during autonomous operation in SI-TL are depicted in Figs.~\ref{fig:fig13} and~\ref{fig:fig14}. Owing to space limitations, the response values for task command ``A'' and the speed commands 3.00, 5.00, and 11.00 are shown. As indicated in the figure, SI-TL could change behavior in the time direction according to the completion time command. This behavior was not simply a linear expansion, but the force at each joint was appropriately modified according to the speed and relationship between each joint. 
By learning multiple motions relating to the speed, SI-TL could learn physical phenomena that were strongly related to the speed and posture through the robot body motion. 
\section{CONCLUSIONS}
\label{sec:sec6}
In this paper, we proposed a variable speed motion generation method for multiple actions. First, we considered models that learned the completion time and task commands. Among the four models examined, only SI-TL, which inputs the completion time command and robot response value into the input layer, and the task command into the final layer of the LSTM, could generate variable speed motions of multiple actions with a high success rate. We also used this model to verify the reproducibility of the commands and found that it could reproduce unlearned speed commands appropriately.

It has recently been reported that the performance of models can be improved by training them based on subtasks different from the target task in the field of self-supervised learning~\cite{c28}. As the task completion time is an additional task that can be provided for almost any task and it can be treated as a subtask in self-supervised learning, self-supervised learning can be achieved in almost any task using the proposed framework. It has the potential to improve the performance of the model by learning the dynamics of itself and its environment more accurately by adding  the task completion time subtasks. Note that self-supervised learning has conventionally been developed for spatial information, and temporal information has not been fully utilized. The proposed framework opens the possibility for self-supervised learning, which significantly contributes to understanding the dynamic phenomena in robotic tasks. In the future, we will examine the effectiveness of the proposed method in more detail by using a robot with multiple DOFs to perform a task that is greatly affected by inertia. Subsequently, we will consider how the task can be executed more accurately within a specified task completion time.


%%%%%%%%%%%%%%%%%%%%%%%%%%%%%%%%%%%%%%%%%%%%%%%%%%%%%%%%%%%%%%%%%%%%%%%%%%%%%%%%

%%%%%%%%%%%%%%%%%%%%%%%%%%%%%%%%%%%%%%%%%%%%%%%%%%%%%%%%%%%%%%%%%%%%%%%%%%%%%%%%

%%%%%%%%%%%%%%%%%%%%%%%%%%%%%%%%%%%%%%%%%%%%%%%%%%%%%%%%%%%%%%%%%%%%%%%%%%%%%%%%
\section*{Acknowledgment}
This research was also supported by the Adaptable and Seamless Technology Transfer Program through Target-driven R\&D (A-STEP) from the JST, Grant Number JPMJTR20RG
%\section*{References}

\end{document}